\documentclass{article}
\usepackage{spconf, amsmath,graphicx}
\usepackage{subcaption}
\usepackage{acronym} 
\usepackage{amsmath}
\usepackage{multirow}
\usepackage[table,xcdraw]{xcolor}
\usepackage{cite}
\usepackage{hyperref}
\hypersetup{
    colorlinks=true,
    linkcolor=blue,
    filecolor=magenta,      
    urlcolor=cyan,
    citecolor=blue
    }

\usepackage[absolute,showboxes]{textpos}

\TPMargin{5pt}

\newcommand{\copyrightstatement}{
    \begin{textblock}{0.84}(0.08,0.93)    
         \noindent
         \footnotesize
         \copyright 2022 IEEE. Personal use of this material is permitted. Permission from IEEE must be obtained for all other uses, in any current or future media, including reprinting/republishing this material for advertising or promotional purposes, creating new collective works, for resale or redistribution to servers or lists, or reuse of any copyrighted component of this work in other works.
    \end{textblock}
}

\title{Transformer Compressed Sensing via Global Image Tokens}
\name{Marlon Bran Lorenzana, Craig Engstrom, and Shekhar S. Chandra}
\address{The University of Queensland, Brisbane, Australia}
\begin{document}
%
\maketitle

\begin{abstract}
	
Convolutional neural networks (CNN) have demonstrated outstanding Compressed Sensing (CS) performance compared to traditional, hand-crafted methods. However, they are broadly limited in terms of generalisability, inductive bias and difficulty to model long distance relationships. Transformer neural networks (TNN) overcome such issues by implementing an attention mechanism designed to capture dependencies between inputs. However, high-resolution tasks typically require vision Transformers (ViT) to decompose an image into patch-based tokens, limiting inputs to inherently local contexts. We propose a novel image decomposition that naturally embeds images into low-resolution inputs. These Kaleidoscope tokens (KD) provide a mechanism for global attention, at the same computational cost as a patch-based approach. To showcase this development, we replace CNN components in a well-known CS-MRI neural network with TNN blocks and demonstrate the improvements afforded by KD. We also propose an ensemble of image tokens, which enhance overall image quality and reduces model size. Supplementary material is available: \href{https://github.com/uqmarlonbran/TCS.git}{https://github.com/uqmarlonbran/TCS.git}.

\end{abstract}
\copyrightstatement
\vspace{5pt}
\section{Introduction}
\label{sec:intro}

\ac{TNN} have been established as the gold-standard for sequence-to-sequence prediction problems, such as natural language processing~\cite{vaswani_attention_2017}. This has been largely attributed to their ability of dynamically adjusting receptive fields, model global dependencies and scale with large amounts of data. The \ac{ViT} succeeded \ac{TNN} contributions by delivering state-of-the-art \ac{CV} performance~\cite{dosovitskiy_image_2020}. \ac{ViT} improved upon \ac{CNN}-based image classification and demonstrated superior scaling with model and dataset size. A current topic of research regarding \ac{ViT} lies in the choice and treatment of the image tokens used as inputs, where naively, patch tokens were initially deployed. Recent work has shown that more efficient image representations can reduce training requirements and overall model size~\cite{yuan_tokens--token_2021, fan_multiscale_2021, liu_swin_2021, ho_axial_2019, xiao_early_2021}. 
\begin{figure}[ht]
	\centering
	\begin{subfigure}{0.47\linewidth}
		\centering
		\includegraphics[width=\linewidth]{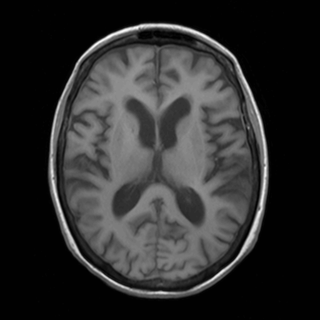}
		\caption{$320\times 320$ sample T1-w brain MRI.}
		\label{fig:kdph}
	\end{subfigure}
	\begin{subfigure}{0.47\linewidth}
		\centering
		\includegraphics[width=\linewidth]{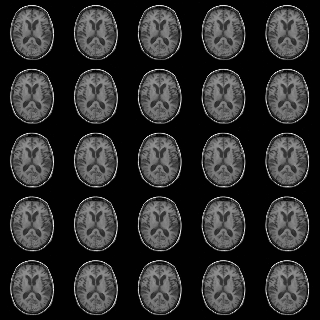}
		\caption{$64\times 64$ Kaleidoscope Embeddings.}
		\label{fig:kde}
	\end{subfigure}
	\caption{$(\nu=5,\sigma=1)$-KT for token embedding: (a) Input Image; (b) Downsampled and concatenated versions of (a).}
	\label{fig:kaleidoscope}
\end{figure}

More specifically, image patches alone present a high level of difficulty for capturing global structures such as edges and lines~\cite{yuan_tokens--token_2021}, and restrict the \ac{ViT} to a single scale of vision~\cite{yuan_tokens--token_2021, fan_multiscale_2021}. What followed was a paradigm of hierarchical \ac{ViT} architectures that aggregate and combine patches in a manner that reduces overall \textit{spatial} resolution and token size~\cite{yuan_tokens--token_2021, fan_multiscale_2021, liu_swin_2021}. The objective being to induce an architectural \textit{prior} that enforces earlier layers to operate on high spatial resolutions, and limit deeper layers to spatially coarse, complex features. While such approaches are useful if dimensionality reduction is desired, it may not be ideal if the output dimension is the same as that of the input. As an alternative to patches, axial tokens have shown great promise for \ac{CV} tasks~\cite{ho_axial_2019}, wherein attention between rows and columns are evaluated independently. Axial attention can be considered a natural extension of \ac{TNN} to \ac{2D} data, akin to decomposing the \ac{DFT} of a \ac{2D} image into a series of \ac{1D} \ac{DFT} operations. Axial encoders are primarily handicapped by necessitating two attention blocks per-encoder, double the feed-forward layers, and a bias towards frequency content along the axes. Lastly, it has been discovered that the inherent local bias' of \ac{CNN} can help improve training stability and overall \ac{CV} performance~\cite{xiao_early_2021}. The central premise being that \ac{CNN}s are able to perform efficient feature extraction, allowing \ac{ViT} to make decisions based on these curated inputs. 

These advancements within \ac{CV} have led to active development of \ac{ViT}s for medical imaging tasks, particularly medical image segmentation~\cite{valanarasu_medical_2021}. However, despite the advantages compared to \ac{CNN} equivalents, they have not yet received significant attention for \ac{CS}. To that end, we propose a \ac{ViT}-based \ac{CS} network for \ac{CS}-\ac{MRI}, which to our knowledge, provides the first convolution-free \ac{DNN} that achieves state-of-the-art reconstruction performance. This paper develops the following:
\begin{enumerate}
    \item Novel \ac{KD} that present global image contexts with low-frequency features; improves overall performance compared to patch alternative.
    \item A cascaded \ac{ViT} architecture that separately attends to image features by employing an ensemble of token embeddings.
    \item Efficient multi-scale \ac{ViT} perception without reducing input or image resolution.
\end{enumerate}
The proposed network architecture extends the \ac{DcCNN}~\cite{schlemper_deep_2017}, wherein \ac{CNN} components are replaced with \ac{TNN} encoder layers. This paper experiments with patch, axial, and our novel \ac{KD} embeddings (see Fig.~\ref{fig:kaleidoscope}). We demonstrate that a \ac{KD}-based \ac{ViT} provides an inherently global context of an image, which when cascaded with patch and axial \ac{ViT} ensures both global and local features are processed independently. This ensemble of \ac{ViT} is an efficient method of achieving multi-scale perception, as the separate features are attended to without affecting resolution and the \ac{KT}~\cite{white_bespoke_2021} needs only the computation cost of ``patchifying" an image.

\section{Method}
\label{sec:method}

\subsection{Kaleidoscope Embedding for Image Tokens}
\label{sec:kd}

\begin{figure}[t]
	\centering
	\begin{subfigure}{\linewidth}
	    \centering
    	\includegraphics[width=\linewidth]{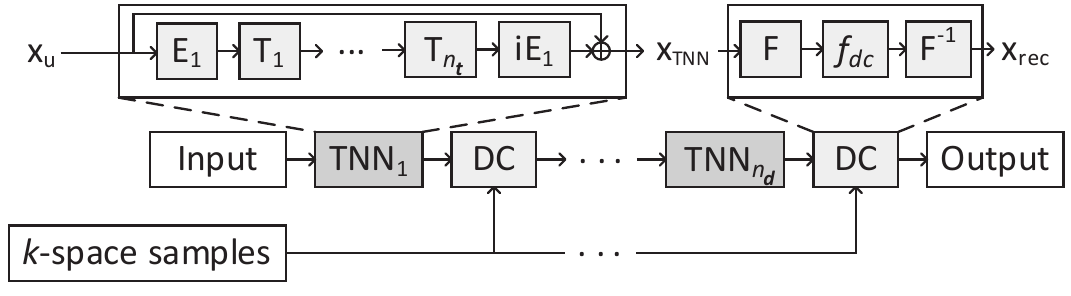}
    	\caption{$\mathbf{E}$ is the token embedding and $i\mathbf{E}$ its inverse, $\mathbf{F}$ is the \ac{DFT}, $n_t$ are the number of transformer encoder layers per-\ac{TNN}, $n_d$ are the number of cascaded \ac{TNN} blocks and $f_{dc}$ is data consistency.}
    	\label{fig:dctnn}
	\end{subfigure}
	\begin{subfigure}{\linewidth}
	    \centering
	    \includegraphics[width=\linewidth]{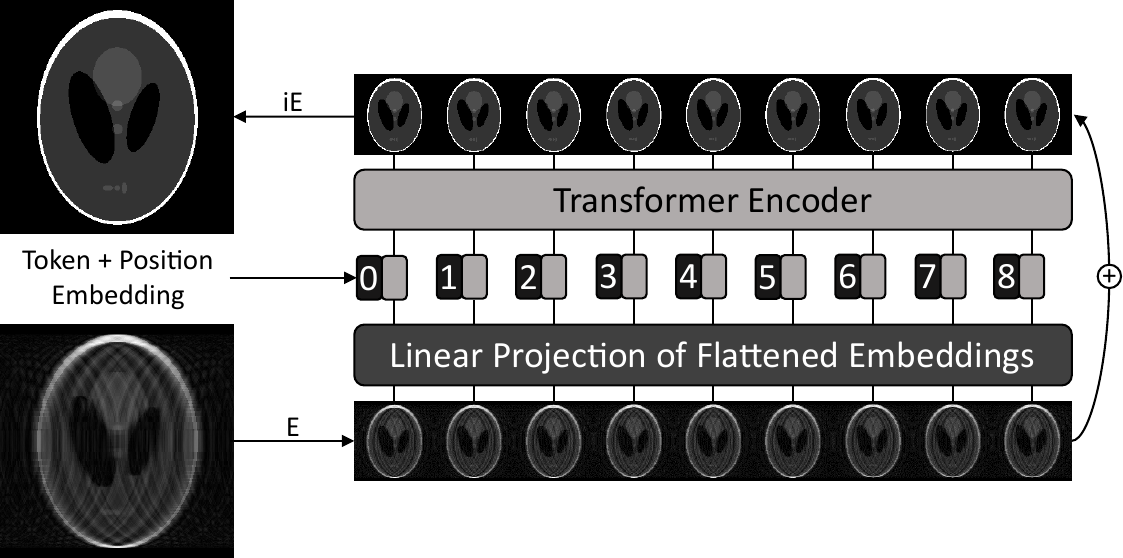}
    	\caption{Example of a \ac{ViT}-based \ac{TNN} denoiser with \ac{KD}. Patch, \ac{KD} and axial tokens are used in this work.}
    	\label{fig:vit}
	\end{subfigure}
	\caption{Architecture for the proposed DcTNN.}
	\label{fig:dctnn-vit}
\end{figure}

The \ac{KT} was recently proposed by White et al.~\cite{white_bespoke_2021} to explain the fractal nature of Chaotic Sensing in \ac{DFT} space~\cite{chandra_chaotic_2018} and formalises the concept of down-sampling and concatenating an image with itself. For example, a $\nu, \sigma$-\ac{KT} decomposes an image into $\nu^2$ down-sampled copies where each is scaled by a factor of $\sigma$. In this work, we efficiently achieve an arbitrary down-sampling factor $\nu$, and smear factor $\sigma = 1$, via element reordering. As seen in Fig. \ref{fig:kaleidoscope} for a $(5, 1)$-KT, $25$ pixel-shifted low-resolution copies of the original image are produced. Image tokens arise via linear projection of the flattened low-resolution images. Fig.~\ref{fig:vit} illustrates its use with \ac{ViT}. A visualisation of exactly how each pixel corresponds to the \ac{KT} is provided in the supplementary material.

Previous \ac{ViT} primarily operate under the assumption that image features can be embedded into patch or axial representations, where a \ac{TNN} encoder is responsible for discerning relationships between these inputs. With respect to image denoising, this assumption dictates that artefacts must be present in a manner that is statistically similar between tokens~\cite{elad_dictionary_2006}. While this is a valid assumption in the context of Gaussian additive noise, \ac{CS} artefacts are not necessarily characterised in such a manner, and therefore vary from token-to-token~\cite{metzler_damp_2016}. Given \ac{KD} are low-resolution copies of the original image, they provide a global image context that ensures structures such as edges, lines and \ac{CS} artefacts can be directly modelled. We demonstrate an improvement to \ac{CS} performance compared to a patch-based approach, which becomes even more pronounced in an ensemble configuration where various token embeddings contribute uniquely to image denoising.

\begin{figure*}[t]
	\centering
	\includegraphics[width=\linewidth]{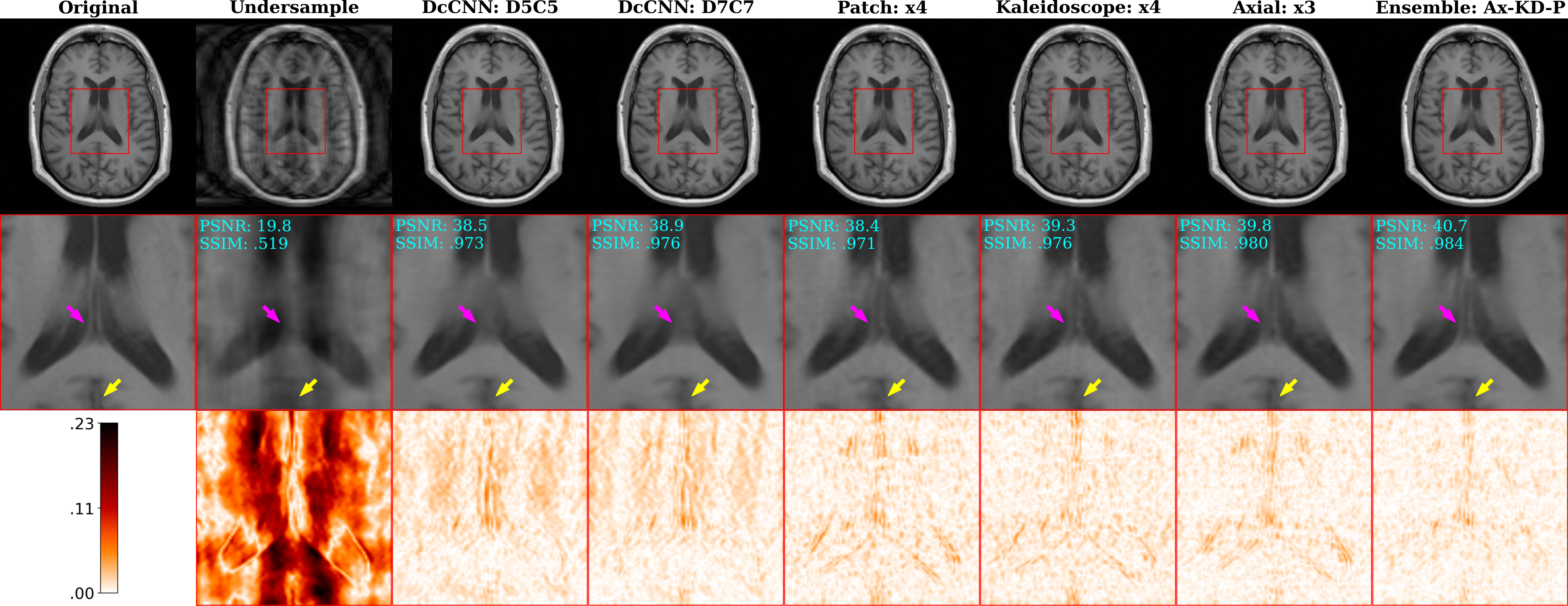}
	\caption{Reconstruction performance demonstrated at R6 reduction factor of a cross-sectional brain MR image, with the zoomed area (red rectangle) including the region of the right and left lateral ventricle. Patch and Kaleidoscope DcTNN are comprised of 4 \ac{TNN} layers. Axial and Ensemble: Ax-KD-P each have 3. Finally, DcCNN are the reference~\cite{schlemper_deep_2017}. Values $\in [0, 1]$.}
	\label{fig:comparisonR6}
\end{figure*}

\subsection{Network Architecture}
\label{sec:netarchi}

In order to effectively deploy \ac{ViT} for \ac{CS}-\ac{MRI}, we impose two major architectural considerations. Firstly, \ac{TNN} blocks should be introduced in a cascaded manner. Secondly, efficient gradient back-propagation is required given the large number of parameters in \ac{TNN}. This study extends on the \ac{DcCNN} as it comprises a relatively simple network architecture that is suitable to these conditions (see Fig.~\ref{fig:dctnn}). \ac{DcCNN} solves the following \ac{CS} optimisation problem,
\begin{align}
    \label{eq:dlmri}
    \min_\mathbf{x,\theta} \; \|\mathbf{x} - f(\mathbf{x})\|^2_2 + \lambda \|\mathbf{\text{F}}_u\mathbf{x} - \mathbf{y}\|^2_2,
\end{align}
where $\mathbf{x}$ is the predicted image, $\mathbf{y}$ is collected discrete Fourier samples, $\mathbf{F}_u$ is the Fourier under-sampling operator, and $f$ regularises the solution. When solved analytically, image estimates $\mathbf{\hat{x}}$ are produced by iteratively applying and adapting $f$ during the reconstruction process~\cite{ravishankar_mr_2011}. \ac{DcCNN} instead replaces $f$ with $f_{cnn, \mathbf{\theta}}$, therefore learning an optimal denoising between iterations in a deep manner; $\mathbf{\theta}$ are the \ac{CNN} parameters. As has been discussed, \ac{CNN}s suffer from inherent limitations that have been overcome by \ac{TNN}s in various applications. We therefore postulate that replacing $f_{cnn,\theta}$ with $f_{tnn,\theta}$ will result in superior image quality, with performance that better scales with available data and total number of parameters. Similarly to \ac{DcCNN}, this \ac{DcTNN} features denoising (\ac{TNN}) and \ac{DC} blocks. \newline

\noindent
\textbf{TNN Denoiser:} \ac{DcCNN} utilises \ac{CNN} denoiser blocks that feature a number of convolutional layers $n_c$, and filters $n_f$. Instead, we employ \ac{TNN} blocks with $n_t$ transformer encoder layers, as-well as an embedding layer $\mathbf{E}$. This embedding can either be common between \ac{TNN} layers, or unique for each. In this way, arbitrary token embeddings such the proposed \ac{KD} can be utilised for \ac{ViT} (see Fig.~\ref{fig:vit}). Our implementation also uses learned positional embeddings. 

We produce a multi-scale perceptive \ac{ViT} by cascading \ac{TNN} denoiser blocks with different token embedding types. In this ``ensemble" design, each \ac{TNN} block can specialise on the type of denoising based on its input features. The ensemble network cascades Axial (Ax), Kaleidoscope (KD), and Patch (P) \ac{TNN} denoising blocks, with Table~\ref{tab:scale} summarising the image features each token attends to. We demonstrate the relative performance improvements afforded by combining these tokens for \ac{CS}-\ac{MRI} in Table~\ref{tab:params} and Fig.~\ref{fig:comparisonR6}. \newline

\noindent
\textbf{Data Consistency:} We implement \ac{DC} blocks that execute:
\begin{equation}
  f_{dc}(\mathbf{\hat{X}}, \mathbf{y}, \lambda) =
    \begin{cases}
      \mathbf{\hat{X}}(k) & \text{if $k \notin \Omega$}\\
      \frac{\mathbf{\hat{X}}(k) + \lambda \mathbf{y}(k)}{1 + \lambda} & \text{if $k \in \Omega$}
    \end{cases}       
\end{equation}
Here $\mathbf{\hat{X}}$ are the discrete Fourier coefficients of the current image estimate $\mathbf{\hat{x}}$, $\lambda$ is a weighting parameter and $\Omega$ represents the subset of sampled points $\mathbf{y}$ with $k$ being an index. In the noiseless case, $\lambda \rightarrow \infty$, and $k \in \Omega$ are directly replaced by $\mathbf{y}$. In our testing, we found that setting $\lambda$ as a learnable parameter in each \ac{DC} block improved performance for \ac{DcTNN}. However, \ac{DcCNN} performed best in the noiseless case. 

\begin{table}[t]
    \centering
    \caption{Relative scales of Vision.}
    \begin{tabular}{l||l|l}
    \textbf{Token}      & \textbf{Image Scale}  & \textbf{Frequency Scale}  \\\hline\hline
    Axial               & Local                 & Low, High                 \\
    KD                  & Global                & Low                       \\
    Patch               & Local                 & High                      \\
    Ensemble: Ax-KD-P   & Local, Global         & Low, High
    \end{tabular}
    \label{tab:scale}
\end{table}

\begin{table*}[t]
\centering
\caption{Average reconstruction performance, number of parameters for tested networks and associated training information. \textbf{Bold} and \underline{underline} indicate best and second best outcomes respectively.}
\begin{tabular}{cclcclcclcclcc}
\hline
                                  & \textbf{Denoising} &  & \multicolumn{2}{c}{\textbf{R=4}}    &  & \multicolumn{2}{c}{\textbf{R=6}}    &  & \multicolumn{2}{c}{\textbf{R=8}}    &  & \textbf{Number of}           & \textbf{Minutes}           \\ \cline{4-5} \cline{7-8} \cline{10-11}
\multirow{-2}{*}{\textbf{Method}} & \textbf{Blocks}    &  & PSNR                         & SSIM &  & PSNR                         & SSIM &  & PSNR                         & SSIM &  & \textbf{Parameters}          & \textbf{Per-Epoch}         \\ \hline \hline
D5C5~\cite{schlemper_deep_2017}   & 5                  &  & \underline{45.61}       & 0.99 &  & 41.31                        & 0.97 &  & 37.68                        & 0.96 &  & \textbf{0.14M} & 14.7                       \\
D7C7~\cite{schlemper_deep_2017}   & 7                  &  & \textbf{46.47}               & 0.99 &  & 41.94                        & 0.98 &  & 37.85                        & 0.96 &  & \underline{1.3M}  & 91.8                       \\ \hline
Patch                             & 4                  &  & 42.68                        & 0.98 &  & 41.43                        & 0.98 &  & 37.31                        & 0.96 &  & 29.3M                        & 5.8                        \\
Kaleidoscope                      & 4                  &  & 43.48                        & 0.98 &  & 41.84                        & 0.98 &  & 37.54                        & 0.96 &  & 29.3M                        & 5.8                        \\
Axial                             & 3                  &  & 44.75                        & 0.99 &  & \underline{42.30}       & 0.98 &  & \underline{38.19}       & 0.97 &  & 38.5M                        & \underline{5.5} \\
Ensemble                          & 3                  &  & 45.01                        & 0.99 &  & \textbf{42.99}               & 0.98 &  & \textbf{38.55}               & 0.97 &  & 27.5M                        & \textbf{4.6} \\ \hline

\end{tabular}
\label{tab:params}
\end{table*}

\section{Results and Discussion}
\label{sec:resdis}

\textbf{Experimental Configuration:} A subset of the NYU fastMRI DICOM brain database was used to train and test models, constituting $64,180$ T1-w cross-sectional MR images~\cite{zbontar2019fastmri}. Training, validation and testing consisted of 80\%, 10\% and 10\% of these images. We simulate single-coil magnitude images and cropped each to $320\times320$ resolution. Discrete Fourier space was sampled using a \ac{1D} Gaussian random mask. Closeness to the original image is measured with \ac{PSNR} and \ac{SSIM}. \ac{DcTNN} networks employ $n_t=2$, where patch and \ac{KD} are $16\times16$ and $d_{model}=256$; $d_{model}=320$ for axial tokens. D5C5 \ac{DcCNN} features $5$ \ac{CNN} blocks with $n_c=5$ and $n_f=32$. D7C7 instead has $7$ \ac{CNN} blocks with $n_c=7$ and $n_f=64$. Further details regarding the training and testing methodology are included in the supplementary material.

\noindent
\textbf{Qualitative Analysis:} Fig.~\ref{fig:comparisonR6} illustrates the reconstruction characteristics of both \ac{DcTNN} and \ac{DcCNN} methods. The arrows point to regions of faint, noise-like image features that all \ac{DcTNN} are able to recover; \ac{DcCNN} over-smooths such regions. Although individual patch, \ac{KD} and axial \ac{DcTNN} leave similar residual artefacts, ensemble (Ax-KD-P) is capable of accurately discerning the underlying image. Additionally, the ensemble network is more faithful to the original image textures compared to \ac{DcCNN}. These reconstruction characteristics are persistent for all reduction factors tested. In fact, despite \ac{DcCNN} scoring higher \ac{PSNR} and \ac{SSIM} at R4 (Table~\ref{tab:params}), image features such as those portrayed in Fig.~\ref{fig:comparisonR6} still lack significant recovery. Visual comparisons at R4 and R8 are available in the supplementary material.\newline

\noindent
\textbf{Quantitative Analysis:} We evaluated the reconstruction performance of our \ac{DcTNN} at several under-sampling rates and compared the results to \ac{DcCNN} in Table~\ref{tab:params}. We found that a \ac{DcTNN} composed entirely of \ac{KD} embeddings outperforms an equivalent patch-based approach, behaving similarly to the axial case; the axial network required more parameters. Importantly, the ensemble of cascaded axial, \ac{KD} and patch \ac{TNN} notably reduces the number of parameters required whilst achieving the highest \ac{PSNR} and \ac{SSIM} scores. Further, training time per-epoch is just a fraction of that required for \ac{CNN} alternatives. These results indicate that our ensemble \ac{DcTNN} is most capable of recovering image information at higher reduction factors, i.e., R6 and R8. It also highlights that the combination of token embeddings can be effectively leveraged for multi-scale processing, resulting in smaller and more efficient networks.

\begin{figure}[t]
	\centering
	\includegraphics[width=\linewidth]{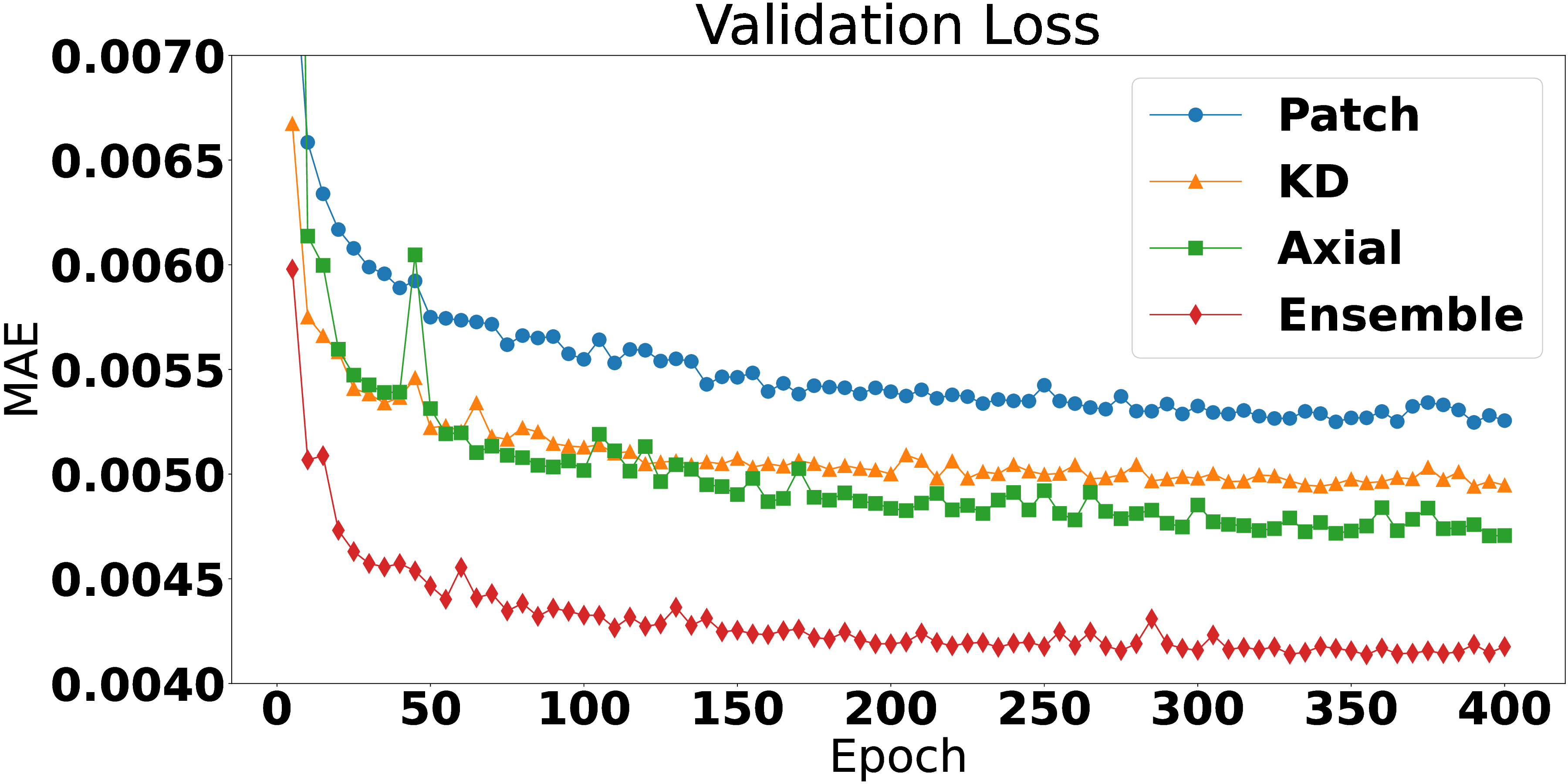}
	\caption{\ac{DcTNN} validation loss (\ac{MAE}) at R6. Ensemble demonstrates superior convergence.}
	\label{fig:valloss}
\end{figure}

Fig.~\ref{fig:valloss} illustrates the comparative training characteristics of the \ac{DcTNN} methods. Here, we see the naive patch-based approach requires many epochs for convergence. At the same number of parameters, \ac{KD} is able to converge faster, and to a lower loss value. While the axial network requires more parameters, overall convergence is similar to that attained with \ac{KD}. Finally, the ensemble network achieves the lowest overall validation loss, as-well as the fastest convergence. These findings demonstrate that \ac{KD} surpasses image patches at efficiently modelling input characteristics. The ensemble network further benefits from the many image scales presented as inputs throughout reconstruction.

\section{Conclusion}
\label{sec:conclusion}

This paper has demonstrated the benefits afforded by our novel Kaleidoscope token (KD) embeddings, which provides low frequency, global image representations. Their use for \ac{CS}-\ac{MRI} improved training and reconstruction characteristics compared to image patches due to superior modelling of global features such as \ac{CS} artefacts. We also present the advantages of cascading token types, therefore encouraging \ac{TNN} layers to specialise at ``seeing" particular image features. This combined approach ensures a multi-scale perceptive \ac{ViT} without a reduction of resolution, or necessitating a windowed approach that may reduce overall vision scales. 

Future work should investigate convolutions within \ac{DcTNN}, as \ac{CNN} are known to improve training characteristics of \ac{ViT}. We also theorise that the total number of parameters can be reduced via the \ac{KT}. In this proposed configuration, low-resolution copies of the image can be attended to separately without reducing vision scales. In conclusion, the advantages of utilising more than a single form of token embedding has been demonstrated, with our novel \ac{KD} producing promising results. We propose that similar tokenisation can be successfully employed in many other fields of \ac{CV}.

\vfill
\pagebreak
\newpage

\bibliographystyle{IEEEbib}
\bibliography{transformercs.bib}

\acrodef{MRI}{magnetic resonance imaging}
\acrodef{MR}{magnetic resonance}
\acrodef{CS}{Compressed Sensing}
\acrodef{TNN}{Transformer neural networks}
\acrodef{ViT}{Vision Transformer}
\acrodef{CV}{computer vision}
\acrodef{NLP}{natural language processing}
\acrodef{DNN}{deep neural network}
\acrodef{CNN}{convolutional neural network}
\acrodef{1D}{one-dimensional}
\acrodef{2D}{two-dimensional}
\acrodef{FFT}{fast Fourier transform}
\acrodef{DFT}{discrete Fourier transform}
\acrodef{KD}{Kaleidoscope tokens}
\acrodef{KT}{Kaleidoscope transform}
\acrodef{DcCNN}{Deep cascade of Convolutional Neural Networks}
\acrodef{DcTNN}{Deep cascade of Transformer Neural Networks}
\acrodef{DC}{data consistency}
\acrodef{ChaoS}{Chaotic Sensing}
\acrodef{DC}{data consistency}
\acrodef{MAE}{mean-absolute-error}
\acrodef{MSE}{mean-squared-error}
\acrodef{TV}{total-variation}
\acrodef{PSNR}{peak signal-to-noise ratio}
\acrodef{SSIM}{structural similarity}
\acrodef{GPU}{graphics processing unit}
\end{document}